\documentclass[letterpaper, 10pt, conference]{ieeeconf}
\IEEEoverridecommandlockouts
\usepackage{amsmath}
\usepackage{amssymb}
\usepackage{cite}
\usepackage{graphicx}
\usepackage{siunitx}
\makeatletter
\let\MYcaption\@makecaption
\makeatother
\usepackage[font=footnotesize]{subcaption}
\makeatletter
\let\@makecaption\MYcaption
\makeatother
\usepackage{pgfplots}
\usepackage{url}
\usepackage[T1]{fontenc}
\usepackage{inconsolata}
\usepackage{mathtools}
\usepackage{lipsum}
\usepackage{booktabs}
\usepackage{makecell}

\let\vec\boldsymbol

\newcommand{\conflobs}{\vec{o}}
\newcommand{\mlpgap}{\text{MLP}_\text{gap}}
\newcommand{\mlpacc}{\text{MLP}_\text{acc}}

\newcommand{\gapaccij}{\delta_{i,j}}
\newcommand{\gapaccigt}{\delta_{i,\text{GT}}}
\newcommand{\veh}{\nu}
\newcommand{\fms}{ms^{-1}}
\newcommand{\fmss}{ms^{-2}}
\usepackage{hyphenat}
\title{\LARGE \bf
Fast Long-Term Multi-Scenario Prediction \linebreak
for Maneuver Planning at Unsignalized Intersections
}
\author{Max Bastian Mertens, Jona Ruof, Jan Strohbeck, and Michael Buchholz%
	\thanks{This work was financially supported by the Federal Ministry for Economic Affairs and Climate Action of Germany within the program ``Highly and Fully Automated Driving in Demanding Driving Situations'' (project LUKAS, grant number 19A20004F).}%
	\thanks{The authors are with the Institute of Measurement, Control and Microtechnology,
		Ulm University, D-89081 Ulm, Germany.
		E-mail addresses: {\tt firstname.lastname@uni-ulm.de}
	}%
}
\newcommand\copyrighttext{%
\footnotesize Copyright $\copyright$ 2024 AACC.
Personal use of this material is permitted.
Permission from AACC must be obtained for all other uses, in any current or future media, including reprinting/republishing this material for advertising or promotional purposes, creating new collective works, for resale or redistribution to servers or lists, or reuse of any copyrighted component of this work in other works.}%
\newcommand\copyrightnotice{%
\begin{tikzpicture}[remember picture,overlay]%
	\node[anchor=south,yshift=10pt] at (current page.south) {\fbox{\parbox{\dimexpr\textwidth-2cm}{\copyrighttext}}};%
\end{tikzpicture}%
\vspace{-10pt}%
}
\begin{document}
\maketitle
\copyrightnotice
\thispagestyle{empty}
\pagestyle{empty}

\begin{abstract}
Motion prediction for intelligent vehicles typically focuses on estimating the most probable future evolutions of a traffic scenario.
Estimating the gap acceptance, i.e., whether a vehicle merges or crosses before another vehicle with the right of way, is often handled implicitly in the prediction.
However, an infrastructure-based maneuver planning can assign artificial priorities between cooperative vehicles, so it needs to evaluate many more potential scenarios.
Additionally, the prediction horizon has to be long enough to assess the impact of a maneuver.
We, therefore, present a novel long-term prediction approach handling the gap acceptance estimation and the velocity prediction in two separate stages.
Thereby, the behavior of regular vehicles as well as priority assignments of cooperative vehicles can be considered.
We train both stages on real-world traffic observations to achieve realistic prediction results.
Our method has a competitive accuracy and is fast enough to predict a multitude of scenarios in a short time, making it suitable to be used in a maneuver planning framework.
\end{abstract}

\section{Introduction}
\label{sec:intro}
\begin{figure}
	\centering
	\includegraphics[width=0.99\linewidth]{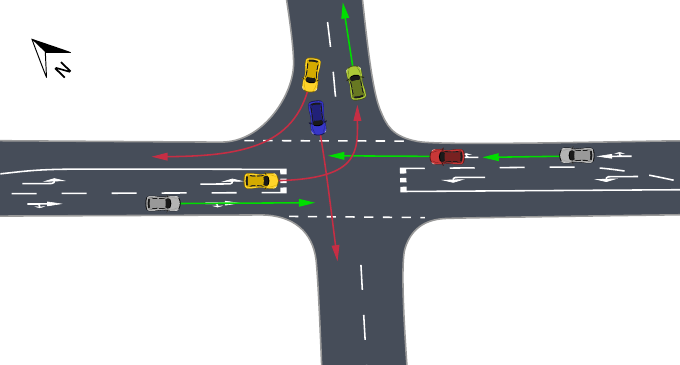}
	\caption{%
		Unsignalized intersection near Bendplatz in Aachen, Germany, and a traffic scenario with future vehicle paths, as found in the inD dataset \cite{bock_ind_2020}.
		Vehicles with red arrows have to yield.
		This scenario has 8 pairs of conflicting vehicles and 14 possible outcomes, i.e. vehicle crossing orders.
	}
	\label{fig:sample_scenario}
\end{figure}

Vehicle-to-vehicle and vehicle-to-infrastructure communication is increasingly widespread in everyday traffic.
Cooperative vehicles can share information about their perception and planned route leading to higher traffic safety \cite{buchholz_handling_2021}.
At unsignalized intersections, like the one shown in Fig.~\ref{fig:sample_scenario}, modern traffic control approaches like infrastructure\hyp{}supported maneuver planning modules can be deployed.
They assign an artificial priority between cooperative vehicles, temporarily deviating from usual right-of-way regulations when it can be coordinated safely.
Cooperative maneuvers can significantly increase the traffic throughput and lower waiting times, depending on the number of cooperative vehicles \cite{dresner_multiagent_2004,mertens_cooperative_2022,klimke_automatic_2023}.
However, traffic control systems depend on a realistic long-term (i.e., $\ge\SI{10}{s}$) prediction of the road users to correctly estimate the efficiency and safety impact of a maneuver.

In early individual traffic control approaches, all vehicles are assumed to be cooperative and automated, so a simple prediction model is sufficient without compromising on safety \cite{dresner_multiagent_2004}.
However, in the medium term, we will still face mixed traffic with human drivers, whose behavior cannot be controlled and has to be predicted \cite{mertens_cooperative_2022,klimke_automatic_2023}.
This is challenging, since not only the continuous velocity of a vehicle has to be estimated, but also the discrete decision of whether it will cross or merge onto another lane, commonly referred to as \emph{gap acceptance}, has to be predicted.

When evaluating potential maneuvers in a planning module, multiple scenarios have to be predicted and compared to the default case with no intervention \cite{mertens_cooperative_2022}.
Therefore, the prediction needs to consider the assigned priorities between road users in a maneuver.
Multiple trajectory prediction approaches like \cite{strohbeck_multiple_2020} might seem suitable to be used in such maneuver planning.
However, the individual vehicle trajectory hypotheses may not be consistent with each other and no explicit maneuver input can be given.
Additionally, the number of potential maneuvers in a scenario like in Fig.~\ref{fig:sample_scenario} can be large, so a fast multi-scenario prediction is necessary.

Towards these outlined challenges, we suggest a motion prediction approach where we handle the gap acceptance estimation and the velocity prediction in separate modules.
Our method, therefore, supports the consideration of priority assignments resulting from potential maneuvers.
We focus on an efficient implementation with longitudinal predictions along the vehicle path to enable the evaluation of a multitude of long-term future scenarios.

The remainder of this paper is structured as follows:
First, we review existing parametric and machine learning approaches for gap acceptance and velocity prediction in Section~\ref{sec:background}.
In Section~\ref{sec:method}, we present our suggested approach and implementation considerations.
We evaluate our method on a validation dataset in Section~\ref{sec:experiments} and provide conclusions.

\section{Related Work}
\label{sec:background}
As motion predictions are needed in various applications, such as vehicle trajectory planning or infrastructure-based cooperative maneuver planning, many approaches have already been proposed.
In this section, we examine central prediction concepts related to our proposed method.

\subsection{Classic Parametric Driver Models}
A traffic scenario can be predicted by iteratively integrating a \emph{driver model} for each individual vehicle along its lane.
These models commonly describe the acceleration of a single vehicle $\veh_i$ by a relation
\begin{equation}
\label{eq:base_driver_model}
\dot{v}_i(t)=f_{\vec{p}}(\vec{o}_i(t))
\end{equation}
with parameters $\vec{p}$ and observations $\vec{o}_i(t)$ about, e.g., the vehicle state (pose and velocity) and the environment (lanes, traffic rules, and other vehicles).
In classic parametric driver models, the function is manually designed based on observations of real traffic data.
The parameters are adapted to closely imitate realistic driving behavior.
A common model of this form is the Intelligent Driver Model (IDM) \cite{treiber_congested_2000}
\begin{subequations}
\begin{align}
\dot {v}_i = a\,\left(1-\left(\frac {v_i}{v_0}\right)^\delta-\left(\frac {s^{*}(v_i,\Delta v_i)}{s_i}\right)^2\right) \\
\text{with } s^{*}(v_i,\Delta v_i) = s_0+v_i\,T+\frac {v_i\,\Delta v_i}{2\,\sqrt{a\,b}},
\end{align}
\end{subequations}
with distance $s_i$ and relative velocity $\Delta v_i$ to the lead vehicle, target velocity $v_0$, minimum distance $s_0$, safety distance $T$, maximum acceleration $a$ and deceleration $b$, and an acceleration exponent $\delta$.
The IDM has a high velocity prediction accuracy and is often used as a baseline although being less realistic than machine learning-based approaches \cite{lefevre_comparison_2014}.

\subsection{Gap Acceptance}
\label{sec:classic_gap_acc}

Driver models often focus on highway traffic and predict the vehicle velocity depending on only the lead vehicle on their own lane \cite{treiber_congested_2000}.
However, at intersections, vehicles from different lanes approach the same area due to crossing or merging onto another lane.
For each conflicting pair, there is one prioritized and one yielding vehicle, determined by the right-of-way regulation of the intersection.
Vehicles required to yield have to decide whether they accept the gap, i.e., enter the intersection before the oncoming prioritized vehicle, or whether they remain behind their respective stop line.
In turn, prioritized vehicles only need to react to lead vehicles that are already ahead in their lane.

Heuristic models for gap acceptance at stop-controlled intersections have existed for decades \cite{greenshields_traffic_1947}.
They assume a temporal \emph{critical gap} length, above which vehicles from the side road are likely to cross (approx. \SI{6}{s}) or merge (approx. \SI{4}{s}).
The microscopic traffic simulator SUMO can also handle unsignalized intersections without stop signs using a time-based gap length model and calculating collision-free time slots on the intersection for each vehicle \cite{behrisch_sumos_2014}.
The authors of \cite{abhishek_generalized_2019} generalize critical gap approaches and incorporate queuing models to achieve more realistic results and to represent driver impatience.

\subsection{Machine Learning-based Prediction Approaches}

The aforementioned classic driver models and gap acceptance heuristics are manually designed and based on intuitive assumptions from traffic observation.
The equations have a small number of parameters where each term corresponds to a physical relation.
On the other hand, machine learning-based approaches like neural networks make fewer assumptions about the function they approximate and typically have orders of magnitude more parameters than heuristic models.
This allows them to achieve higher accuracies when they are trained on a proper dataset.

Reinforcement Learning (RL) is often used for continuous control or prediction problems \cite{duan_benchmarking_2016}.
RL algorithms like Proximal Policy Optimization (PPO) \cite{schulman_proximal_2017} aim to train a \emph{policy} to achieve high values of a given \emph{reward function} in a simulation environment.
The policy, typically implemented by a neural network, receives an \emph{observation} vector as the input and results in an \emph{action} output.
This resembles the structure of Eq.~\eqref{eq:base_driver_model} and can be evaluated iteratively in the same manner.
Provided a traffic simulation and a reward function that considers traffic rules and penalizes collisions, a driver model policy can be learned.
These models can be used for long-term predictions with a \SI{10}{s} horizon and often implicitly include the gap acceptance modeling \cite{kuefler_imitating_2017,sackmann_modeling_2022}.

While manually engineering the reward function is possible, it can be a tedious process and might not yield realistic behavior in the resulting policy.
Inverse Reinforcement Learning (IRL) methods like Generative Adversarial Imitation Learning (GAIL) \cite{ho_generative_2016} thus try to automatically extract a reward function from a dataset.
They employ another discriminator network that distinguishes states resulting from policy actions and states from expert trajectories.
IRL has been applied to the task of training driver models before \cite{kuefler_imitating_2017,sackmann_modeling_2022}.
Since the training of adversarial methods like GAIL can be unstable, the network can be pre-trained in a simulation environment with a manually defined reward function.

Direct learning approaches
like image-based convolutional neural networks \cite{strohbeck_multiple_2020}
are often used for short-term predictions with a horizon time of $\SI{3}{s}$, but can be evaluated iteratively to reach long-term horizons of \SI{10}{s} \cite{strohbeck_deepsil_2021}, which significantly increases the computation time.

The task of predicting discrete driver decisions can also be performed separately by a machine learning approach.
It can be modeled as a Markov Decision Process (MDP) and solved using 
a combination of RL and Monte Carlo Tree Search \cite{kurzer_decentralized_2018}, returning a probability distribution over possible decisions instead of a single predicted action.
MDP-based methods are, however, computationally very demanding \cite{kurzer_decentralized_2018}, making them unsuitable for our intended application.

\section{Method}
\label{sec:method}

This section describes our problem scope and approach to motion prediction at unsignalized intersections.

\subsection{Problem Definition and Assumptions}
\label{sec:problem_definition}

The goal of our work is to predict the future trajectories, i.e., positions and velocities, of a traffic scenario at an intersection, as depicted in Fig.~\ref{fig:sample_scenario}, over a horizon of \SI{10}{s}.
To accurately estimate the traffic efficiency during maneuver planning, the predicted states should be as close to real traffic as possible.

The current position and velocity of each vehicle are assumed to be known, which can be achieved by an in-car or infrastructure perception system as the one in \cite{buchholz_handling_2021}.
The future path, i.e., turn decisions, is assumed to be known as well, which is a common assumption in driver model approaches \cite{lefevre_comparison_2014,kuefler_imitating_2017,sackmann_modeling_2022}.
Since there are existing approaches predicting the path taken at intersections \cite{strohbeck_multiple_2020} and similar situations like roundabouts \cite{sackmann_classification_2020},
this task is not considered in the remainder of this paper and the prediction is only performed longitudinally along the future vehicle path.
Furthermore, no reaction time or delay and no occlusion are considered.

\subsection{Approach}

\begin{figure}[t]
	\vspace{0.08cm}
	\centering
	\begin{subfigure}{\linewidth}
		\centering
		\includegraphics[width=0.85\linewidth]{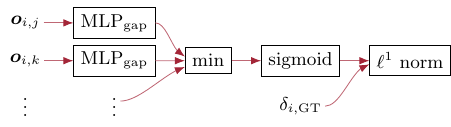}\hspace{0.3cm}
		\caption{%
			Gap acceptance network ($\mlpgap$) architecture during training on the dataset, for vehicle $\veh_i$ with at least one prioritized vehicle $\veh_{j,k,\ldots}$.
		}
		\label{fig:gap_acc_training}
	\end{subfigure}
	\\[0.3cm]
	\begin{subfigure}{\linewidth}
		\centering
		\includegraphics[width=0.99\linewidth]{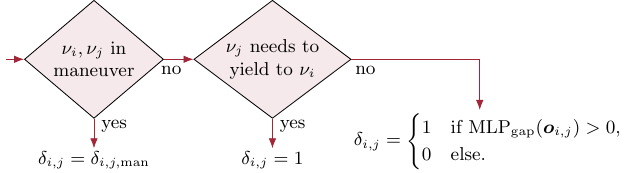}
		\caption{%
			Gap acceptance decision prediction $\gapaccij$ between two vehicles $\veh_i$, $\veh_j$.
		}
		\label{fig:gap_acc_flow}
	\end{subfigure}
	\\[0.3cm]
	\begin{subfigure}{\linewidth}
		\centering
		\includegraphics[width=0.55\linewidth]{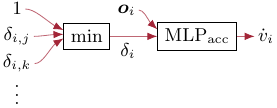}
		\caption{%
			Acceleration prediction for vehicle $\veh_i$ with conflicting vehicles $\veh_{j,k,\ldots}$.
		}
		\label{fig:acc_pred}
	\end{subfigure}
	\caption{Proposed architecture for gap acceptance and motion prediction.}
	\label{fig:prediction_network}
\end{figure}

We employ a dual-stage neural network architecture to solve the task of motion prediction at unsignalized intersections, as depicted in Fig.~\ref{fig:prediction_network}.
The first stage estimates whether a vehicle will enter the intersection.
The second network, the driver model, uses this decision as well as state and environment observations to estimate the acceleration.
Both are evaluated for each vehicle at each integration time step.

\begin{figure}[t]
	\centering
	\def\svgwidth{0.99\linewidth}
	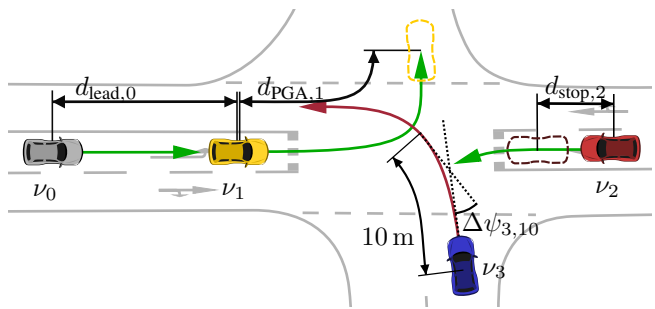
	\caption{%
		Illustration of the observation input features listed in Table~\ref{tbl:agent_observation}.
	}
	\label{fig:agent_observation}
\end{figure}

Using this architecture, a potential maneuver can be predicted by feeding the priority assignment directly to the respective vehicles.
As the maneuver planning needs to predict a large number of possible outcomes over a long-term horizon, the networks need to be evaluated quickly.
Thus, the two stages are implemented using small Multilayer Perceptrons $\mlpgap$ and $\mlpacc$ (cf. Table~\ref{tbl:training_params}), in contrast to other approaches with larger MLPs \cite{kuefler_imitating_2017,sackmann_modeling_2022} or more complex network architectures \cite{strohbeck_multiple_2020}.
The inputs (observations) and outputs (gap acceptance and acceleration) of both stages are depicted in Table~\ref{tbl:agent_observation} and Fig.~\ref{fig:agent_observation} and described hereafter.

\begin{table}[t]
	\vspace{0.2cm}
	\caption{Observation input features of the two prediction MLPs.}
	\label{tbl:agent_observation}
	\begin{center}
		\begin{tabular}{l l}
			\toprule
			\thead[l]{Gap observation $\vec{o}_{i,j}$ \\
				between yielding vehicle $\veh_i$ \\
				and prioritized vehicle $\veh_j$} & \thead[l]{Input features} \\
			\midrule
			Distance to point of guaranteed arrival & $d_{\text{PGA},i}$ \\
			Velocity & $v_i$ \\
			Distance of other vehicle to stop line & $d_{\text{stop},j}$ \\
			Velocity of other vehicle & $v_j$ \\
			\toprule
			\thead[l]{Driving observation $\vec{o}_{i}$ of vehicle $\veh_i$} & \thead[l]{Input features} \\
			\midrule
			Distance to stop line & $d_{\text{stop},i}$ \\
			Velocity & $v_i$ \\
			Current max. velocity & $v_{\max,i}$ \\
			Relative lane heading in $n$ meters & \makecell{$\Delta\psi_{i,-10}$, $\Delta\psi_{i,-3}$, $\Delta\psi_{i,3}$, \\ $\Delta\psi_{i,10}$, $\Delta\psi_{i,30}$, $\Delta\psi_{i,100}$} \\
			Lead vehicle distance & $d_{\text{lead},i}$ \\
			Lead vehicle velocity & $v_{\text{lead},i}$ \\
			\bottomrule
		\end{tabular}
	\end{center}
\end{table}

\subsubsection{Gap Acceptance Prediction}

The task of the first network is to predict the decision $\delta_{i}$ of whether a vehicle $\veh_i$ enters the intersection.
The gap acceptance $\delta_{i,j}(\conflobs_{i,j})$ against every conflicting vehicle $\veh_j$ needs to be predicted, where $\conflobs_{i,j}$ is the gap observation of $\veh_i$.
When a priority $\delta_{i,j,\text{man}}$ is assigned from a maneuver, this value is used instead, as depicted in Fig.~\ref{fig:gap_acc_flow}.
The minimum of all values is $\delta_{i}$.

As explained in Section~\ref{sec:classic_gap_acc}, classical gap acceptance predictions use the temporal gap length of the main road vehicle as an indicator of whether a gap is safe to use.
Therefore, the velocity $v_j$ and distance to the intersection $d_{\text{stop},j}$ of the prioritized vehicle $\veh_j$ are used as input features.

However, fixed critical gap times are only valid for stop-controlled intersections, where yielding vehicles decide about entering with zero velocity and distance to the stop line.
When vehicles approach the intersection from a certain distance and velocity, they might need a smaller or larger time until they did safely cross or merge onto the main road.
Therefore, the velocity $v_i$ and distance to the point of guaranteed arrival (PGA, see Fig.~\ref{fig:agent_observation}) $d_{\text{PGA},i}$ of the yielding vehicle $\veh_i$ are additional input features in $\conflobs_{i,j}$.
The distances $d_{\text{stop},j}$ and $d_{\text{PGA},i}$ are measured along the lane centerlines.

\subsubsection{Driver Model}

The second stage estimates the applied acceleration of a vehicle at each time step along the prediction horizon.
The observation vector comprises information about the vehicle state, road environment, and interaction.
The state of a vehicle $\veh_i$ is given by its distance to the intersection stop line $d_{\text{stop},i}$ and its velocity $v_i$.
The surrounding road environment is represented by the maximum allowed velocity $v_{\max,i}$ at the current position as well as the relative lane heading behind and in front of the vehicle.
The heading values can be used by the network to estimate the local road curvature and lower the velocity accordingly.
Information about a potential lead vehicle is given by the relative distance $d_{\text{lead},i}$ and absolute velocity $v_{\text{lead},i}$ of the closest vehicle in front of $\veh_i$.
If there is no such vehicle, these input features are set to $d_{\text{lead},i}=\SI{100}{m}$ and $v_{\text{lead},i}=\SI{100}{km/h}$, emulating an irrelevant lead vehicle.
Lastly, the gap acceptance decision $\delta_{i}$ is given as an input feature, as shown in Fig.~\ref{fig:acc_pred}.

\subsubsection{Pre-Training with Reward Function}
\label{sec:reward_fun}

\begin{table}[tb]
	\vspace{0.2cm}
	\caption{Penalty terms of the driver model reward function.
	}
	\label{tbl:penalty_terms}
	\begin{center}
		\begin{tabular}{l l l}
			\toprule
			\thead[l]{Penalty condition} & \thead[l]{Violation scaling} & \thead[l]{Factor} \\
			\midrule
			$\left|\dot{v}_i\right| > 0$ & quadratic & $0.01$ \\
			$\left|\dot{v}_i\right| > 2.5$ & constant & $1$ \\
			$\left|\dot{v}_i\right| > 4$ & constant & $20$ \\
			$v_i < v_{\min,i}$ & linear & $20/v_{\max,i}$ \\
			$v_i > v_{\max,i}$ & linear & $30/v_{\max,i}$ \\
			$v_i < 0$ & linear & $30$ \\
			$d_{\text{lead,brake},i} < 0$ & linear & $30$ \\
			$\delta_{i}=0$ and $d_{\text{stop,brake},i} < 0$ & linear & $30$ \\
			\makecell[l]{$d(\vec{x}_i,\vec{x}_j)<\SI{0.5}{m}$ or \\
				$d(\vec{x}_{\text{brake},i},\vec{x}_{\text{brake},j})<\SI{0.5}{m}$} & constant & $5000$ \\
			\bottomrule
		\end{tabular}
	\end{center}
\end{table}

The driver model policy is pre-trained using PPO to stabilize the subsequent IRL training.
We created a multi-agent closed-loop simulation environment and manually designed a reward function to imitate natural driving behavior.
The base reward is $2$ and the penalty terms are given in Table~\ref{tbl:penalty_terms}.
The acceleration is regularized by a small penalty on the quadratic value and by penalizing large values.
The velocity is penalized if $v_i\notin\left[0,v_{\max,i}\right]$.
Upon collisions, the vehicle needing to yield is penalized with a large constant and the simulation is reset.
The distance $d(\vec{x}_i,\vec{x}_j)$ between vehicles $\veh_i,\veh_j$ is calculated by approximating a vehicle bounding box with two circles.
To detect inevitable collisions and the overshooting of stop lines early, full braking positions are introduced:
\begin{subequations}
\begin{align}
	d_{\text{lead,brake},i} &= d_{\text{lead},i} - s_{\text{brake},i}, \\
	d_{\text{stop,brake},i} &= d_{\text{stop},i} - s_{\text{brake},i}, \\
	\text{with } s_{\text{brake},i} &= \tfrac{1}{2}v_i^2/\SI{4}{\fmss}.
\end{align}
\end{subequations}
The vectors $\vec{x}_i$, $\vec{x}_{\text{brake},i}$ describe the current position and the position at distance $s_{\text{brake},i}$ along the route, respectively.

Without intervention, the policy would converge to simply braking until a standstill to prevent collisions.
To avoid this, a minimum required velocity is introduced which is still collision-free even with a conservative deceleration ($\SI{1}{\fmss}$):
\begin{align}
v_{\min,i} &= \min\left\{v_{\max,i},\sqrt{2\cdot \SI{1}{\fmss} \cdot d_{\text{stop},i}^*}\right\}, \\
\text{with } d_{\text{stop},i}^* &= \begin{cases}
	\min\left\{d_{\text{lead},i},d_{\text{stop},i}\right\} &\text{ if } \delta_{i}=0 \\
	d_{\text{lead},i} &\text{ else.}
	\end{cases}
\end{align}
This makes the policy accelerate when possible but does not penalize slowing down for a lead vehicle or stop line.

\section{Experiments}
\label{sec:experiments}

This section describes the implementation details and evaluation results of our approach.

\subsection{Training and Evaluation Dataset}
\label{sec:dataset_preproc}

We selected the inD dataset \cite{bock_ind_2020} comprising road user trajectories at four unsignalized intersections.
We use the traffic observation data from an intersection near Bendplatz in Aachen, Germany, as the urban environment is very similar to the intersection where our prediction and maneuver planning will be deployed \cite{buchholz_handling_2021}.
The dataset was recorded under normal traffic load and in good weather and includes the trajectories of vehicles as well as pedestrians and bicycles.
As our approach considers only vehicles, we exclude scenarios with crossing pedestrians.
However, bicycles on the road have to be considered as leading and conflicting traffic participants.
Therefore, their motion is predicted using the same model, but not considered in the training and evaluation metrics.
As our approach focuses on the longitudinal motion, we project the position and velocity of road users onto their lanes.
Further, we exclude all overtaking scenarios from the dataset, since the projected trajectories would collide.
This results in \SI{1.0}{h} of relevant traffic observations out of the original \SI{3.1}{h} dataset, which is split into a training part (\SI{90}{\percent}) and a validation part (\SI{10}{\percent}).

The traffic observation has a limited field of view, where only \SI{35}{m} of the main road and \SI{20}{m} of the side road are visible on each side of the intersection (see Fig.~\ref{fig:sample_scenario} for an illustration).
At a maximum allowed velocity of \SI{50}{km/h} (\SI{30}{km/h} on the side road), vehicles reach the intersection only about \SI{2.5}{s} after entering the field of view.
This is much lower than a critical gap length of $4-\SI{6}{s}$ (cf. Section~\ref{sec:classic_gap_acc}).
A simple application of driver models would therefore either lead to many collisions or one would have to constantly expect a vehicle entering the observed area on the main road, so there would never be a large enough gap.
Therefore, we extrapolate the trajectories in the dataset over a \SI{6}{s} horizon into the past using a constant velocity model to increase the observed area.
The extrapolated parts of the trajectories are not included in the training and evaluation metrics.

\subsection{Model Training}

\begin{table}[tb]
	\vspace{0.2cm}
	\caption{Training parameters}
	\label{tbl:training_params}
	\begin{center}
		\begin{tabular}{l l}
			\toprule
			\multicolumn{2}{c}{\thead[c]{Gap acceptance network $\mlpgap$ training}} \\
			\midrule
			Training method & supervised learning \\
			Layer sizes & 4\,/\,16\,/\,16\,/\,1 \\
			Activation function & LeakyReLU \\
			Learning rate & $10^{-3}$ (CyclicLR) \\
			Batch size & 8192 samples \\
			Epochs & 1000 \\
			\toprule
			\multicolumn{2}{c}{\thead[c]{Driver model $\mlpacc$ RL pre-training}} \\
			\midrule
			Training method & multi-agent PPO (closed-loop env.) \\
			Layer sizes & 12\,/\,16\,/\,16\,/\,1 \\
			Activation function & $\tanh$ \\
			Learning rate & $2\cdot 10^{-5}$ (constant) \\
			Epoch length & 2000 steps per agent \\
			Epochs & $10^6$ \\
			\toprule
			\multicolumn{2}{c}{\thead[c]{Driver model $\mlpacc$ IRL training}} \\
			\midrule
			Training method & single-agent GAIL (open-loop env.) \\
			Discriminator layer sizes & 12\,/\,64\,/\,64\,/\,1 \\
			Disc. activation function & ReLU \\
			Disc. learning rate & $10^{-3}$ (constant) \\
			Epoch length & 16384 steps \\
			Epochs & 12000 \\
			Policy\,/\,disc. updates per epoch & 10\,/\,4 \\
			\bottomrule
		\end{tabular}
	\end{center}
\end{table}

The two stages of our prediction architecture are trained separately using different machine learning methods.
The training parameters are listed in Table~\ref{tbl:training_params}.

\subsubsection{Gap Acceptance Prediction}

The gap acceptance decision network $\mlpgap$ is trained using supervised learning on scenarios from the dataset.
For each frame, the observation $\conflobs_{i,j}$ between each pair of conflicting yielding vehicle $\veh_i$ and prioritized vehicle $\veh_j$ is recorded as the input values.
Yielding vehicles slowed down by a lead vehicle are excluded from the training data.
The output label is the ground truth flag $\gapaccigt$ of whether $\veh_i$ entered the intersection before all prioritized conflicting vehicles in the observed scenario, as shown in Fig.~\ref{fig:gap_acc_training}.
During training, normally distributed noise is added to the inputs, with $\sigma_{d_{\text{PGA},i}}=\SI{0.5}{m}$, $\sigma_{v,i}=\SI{0.1}{\fms}$, $\sigma_{d_{\text{stop},j}}=\SI{2.0}{m}$, and $\sigma_{v_j}=\SI{2.0}{\fms}$.
This increases the generalization of the network and resembles realistic observation inaccuracies.

As a baseline, we applied SUMO's gap acceptance model (cf. Section~\ref{sec:classic_gap_acc}), assuming constant velocity at $v_{\max}$ to estimate the time until reaching the intersection, which yielded an accuracy of \SI{86.0}{\percent}.
In contrast, the training of our approach converged with an accuracy of \SI{94.5}{\percent} on the validation data, making it a much more realistic prediction.

\subsubsection{Driver Model}

The driver model network $\mlpacc$ is pre-trained using PPO in a multi-agent closed-loop simulation environment.
Throughout model training and evaluation, we use a discrete integration step of \SI{0.2}{s}.
The simulation uses the intersection layout as shown in Fig.~\ref{fig:sample_scenario}, with an extended map of about \SI{150}{m} on the main road and \SI{70}{m} on the side road on each side of the intersection.
At simulation start and reset, 12 vehicles are placed on the map with a random route, position, and velocity such that none of the aforementioned penalty constraints are violated.
All vehicle states are advanced using the same policy and value networks, which are in turn updated based on the accumulated training gradients of all vehicles.
Whenever a collision is detected, the simulation is reset.
When a vehicle leaves the map, it is reinserted at the start of another route.
If this leads to a constraint violation, the simulation is reset entirely.
The training converged to an average reward of $0.4$, less than $10^{-3}$ collisions per second per agent, and an average velocity of \SI{6}{\fms} in a validation simulation of \SI{2000}{s} length.

Afterwards, the policy and value networks are re-trained using GAIL to make the behavior more realistic.
During training, the vehicles are initialized from randomly chosen scenarios in the dataset like the one in Fig.~\ref{fig:sample_scenario}.
Only one vehicle is advanced by the policy, while the other vehicles in the scenario are updated from the dataset, possibly entering or leaving the scenario.
This single-agent open-loop environment forces the policy to behave similarly to the vehicles in the dataset.
The simulation is run until the predicted vehicle or the original vehicle in the dataset leaves the scenario.

\subsection{Evaluation}

\begin{figure}[tb]
	\vspace{0.08cm}
	\centering
	\begin{subfigure}{\linewidth}
		\input{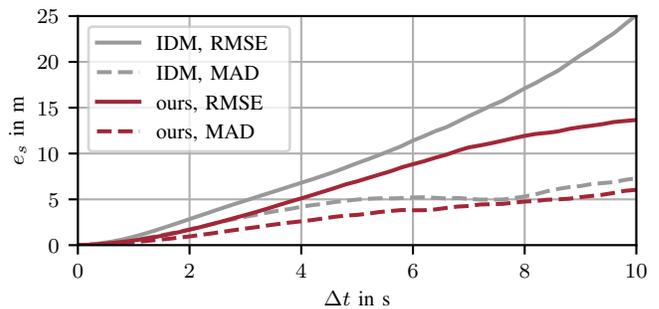}
		\caption[]{Distance error $e_s$ over prediction horizon $\Delta t$.}
		\label{fig:plot_s_over_t}
	\end{subfigure}
	\begin{subfigure}{\linewidth}
		\input{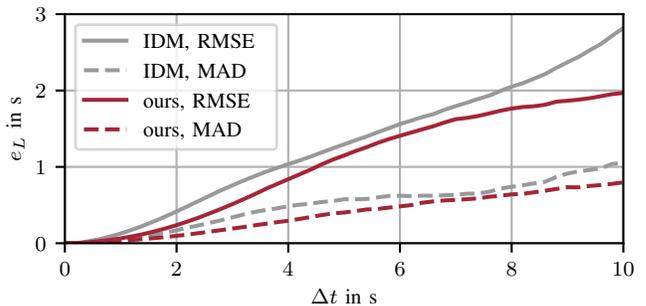}
		\caption[]{Time loss error $e_L$ over prediction horizon $\Delta t$ (cf. Eq.~\eqref{eq:time_loss}).}
		\label{fig:plot_time_loss_over_t}
	\end{subfigure}
	\caption{Prediction accuracy of our $\mlpacc$ vs. IDM on the validation dataset.
		For both, our gap acceptance model was used.
		The root mean square error (RMSE) and median absolute deviation (MAD) per vehicle is shown.}
	\label{fig:plots_eval}
\end{figure}

We evaluated our prediction approach on the traffic scenarios in the validation dataset.
The evaluation results are discussed in the following sections.

\subsubsection{Gap Acceptance Accuracy}

To validate the performance of the gap acceptance module, we predicted the scenarios using our method in the default case, i.e. without giving any maneuver inputs.
We evaluated the predicted crossing order of each conflicting vehicle pair and compared that to the ground-truth order in the validation dataset.
We obtained an accuracy of \SI{83.6}{\percent}, so in most scenarios the gap acceptance is estimated correctly.

\subsubsection{Driver Model Accuracy}

We evaluate the driver model accuracy by comparing the predicted distance along the vehicle path to the ground truth in the validation dataset.
We compare our model against the IDM with parameters chosen similarly to the ones in \cite{kuefler_imitating_2017}:
$s_0=\SI{1.5}{m}$, $T=\SI{1}{s}$, $a=\SI{2.5}{\fmss}$, $b=\SI{4}{\fmss}$, $\delta=4$.
For both models, we used our gap acceptance estimation because the IDM has none.

Figure~\ref{fig:plot_s_over_t} shows the evaluation results.
After a prediction horizon of \SI{10}{s}, our method has a distance root mean square error (RMSE) of \SI{14}{m} (\SI{40}{\percent} less than the IDM) and a median absolute deviation (MAD) of \SI{6}{m} (\SI{20}{\percent} less than the IDM).
Other approaches report comparable results, with an RMSE of \SI{10}{m} (highway scenarios) \cite{kuefler_imitating_2017}, \SI{9}{m} (intersection scenarios) \cite{strohbeck_deepsil_2021}, and \SI{15}{m} (mostly roundabout scenarios) \cite{sackmann_modeling_2022}, respectively, all at the same \SI{10}{s} prediction horizon.
However, they do not support the consideration of priority assignments from potential cooperative maneuvers.

\subsubsection{Multi-Scenario Prediction for Maneuver Planning}

\begin{figure}[tb]
	\vspace{0.08cm}
	\centering
	\input{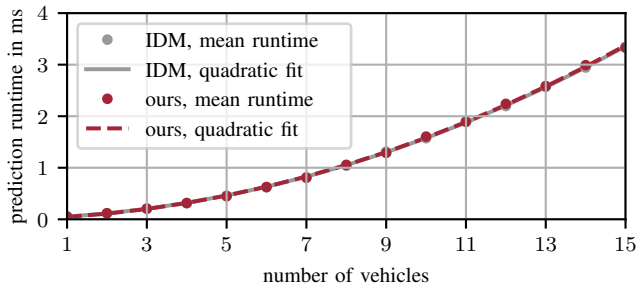}
	\caption[]{Runtimes for a \SI{10}{s} prediction horizon at typical numbers of vehicles in the intersection scenario, averaged over 2000 random scenes.}
	\label{fig:plot_runtime}
\end{figure}

For our maneuver planning approach, we introduced the time loss $L(\veh_i)$ of a vehicle $\veh_i$ as a maneuver efficiency metric \cite{mertens_cooperative_2022}:
\begin{align}
	\label{eq:time_loss}
	L(\veh_i) &\coloneqq \int_{T_\text{start}}^{T_\text{end}} 1-\frac{v_i(t)}{v_{\max,i}(t)} \,\text dt.
\end{align}
In Fig.~\ref{fig:plot_time_loss_over_t}, we evaluated the precision of the predicted time loss of a scenario.
On the validation dataset, our estimation of the vehicle time loss has an error of \SI{2.0}{s} (RMSE) or \SI{2.3}{s} (\SI{80}{\percent} quantile) at the end of the prediction horizon.
This deviation is low compared to a possible average time loss reduction of up to \SI{15}{s} in dense traffic \cite{mertens_cooperative_2022}.
As a result, when our prediction reports a time loss reduction of more than \SI{2}{s} due to a potential maneuver, the real time loss is likely reduced and the maneuver should therefore be executed.

The runtime of our driver model is similar to the IDM and scales quadratically with the number of vehicles due to pairwise interactions (see Fig.~\ref{fig:plot_runtime}).
Because of an efficient C++ implementation, the runtime remains below \SI{3.5}{ms} on a single CPU core (AMD Ryzen 7 3700X) for 15 vehicles at a \SI{10}{s} prediction horizon.
This enables the evaluation of over 50 potential maneuvers in a \SI{5}{\hertz} cycle, making it suitable for maneuver planning.
Most authors of related work do not report on the runtime.
However, in \cite{strohbeck_multiple_2020}, an inference time of \SI{2}{ms} on an NVIDIA Geforce 1080 Ti GPU for a \SI{3}{s} horizon and 6 predicted trajectories per vehicle is stated, being in the same order of magnitude but requiring more resources.

We employed our approach in the cooperative maneuver planning in the LUKAS project \cite{lukas2023video} and demonstrated that it is real-time capable and can help to improve traffic efficiency.

\section{Conclusion}
\label{sec:conclusion}

We presented a new approach to gap acceptance and vehicle motion prediction at intersections.
We achieve similar accuracy as state-of-the-art approaches while considering priority assignments from potential maneuvers.
Our method is able to quickly predict a large number of future scenarios over a long-term horizon, making it suitable to be used in a cooperative maneuver planning environment.
We plan to extend our approach to handle inaccuracies of infrastructure perception.
Furthermore, we want to investigate the prediction of several scenarios per maneuver and their probabilities.

\bibliographystyle{IEEEtran}
{\footnotesize
	\bibliography{root}}
\end{document}